# Uncertainty in AI: Evaluating Deep Neural Networks on Out-of-Distribution Images

Jamiu Adekunle Idowu[1]
University College London
ucabaid@ucl.ac.uk

Ahmed Almasoud[2]
Prince Sultan University
aalmasoud@psu.edu.sa

## Abstract

As AI models are increasingly deployed in critical applications, ensuring the consistent performance of models when exposed to unusual situations such as out-of-distribution (OOD) or perturbed data, is important. Therefore, this paper investigates the uncertainty of various deep neural networks, including ResNet-50, VGG16, DenseNet121, AlexNet, and GoogleNet, when dealing with such data. Our approach includes three experiments. First, we used the pretrained models to classify OOD images generated via DALL-E to assess their performance. Second, we built an ensemble from the models' predictions using probabilistic averaging for consensus due to its advantages over plurality or majority voting. The ensemble's uncertainty was quantified using average probabilities, variance, and entropy metrics. Our results showed that while ResNet-50 was the most accurate single model for OOD images, the ensemble performed even better, correctly classifying all images. Third, we tested model robustness by adding perturbations (filters, rotations, etc.) to new epistemic images from DALL-E or real-world captures. ResNet-50 was chosen for this being the best performing model. While it classified 4 out of 5 unperturbed images correctly, it misclassified all of them post-perturbation, indicating a significant vulnerability. These misclassifications, which are clear to human observers, highlight AI models' limitations. Using saliency maps, we identified regions of the images that the model considered important for their decisions.

## 1.0 Introduction

Artificial intelligence (AI), especially deep neural networks (DNNs), has recorded significant growth in recent years. These tools, which mimic how our brains work, are now used in many applications from detecting tumors in medical images to analyzing satellite imagery for flood assessment, and facial recognition for security systems. But as AI use grows, so do our questions about it. A major concern is how these AI systems behave when they face "unusual situations" - circumstances where data deviates from the norm or has been intentionally altered.

Out-of-distribution (OOD) data challenges our understanding of model reliability. It encompasses data samples that, while possibly related to the training set, exhibit unexpected variations or appear in unfamiliar contexts. When confronted with OOD data, models can make high-confidence yet incorrect predictions, often resulting in potentially risky or misinformed decisions [1]. Moreover, as AI is deeply integrated into applications with substantial real-world implications, the stakes become higher.

In addition to OOD data, perturbed data represents another vital aspect. Perturbations can arise from different sources, be it environmental changes, intentional adversarial attacks, or inherent variations in the data capturing mechanism. It's crucial to assess how models react, adapt, and sometimes fail under these conditions [2].

Therefore, this research investigates the robustness of DNN architectures when exposed to both OOD and perturbed data. Through this, we aim to understand the inherent limitations and potential improvement areas for AI models in dealing with unusual or modified data. Furthermore, we believe that understanding the uncertainty and robustness of these models can pave the way for safer and more reliable AI applications in the future.

## 2.0    Related Work

A common thread in recent literature is the centrality of trust in the outputs provided by AI models. As noted by [3], while in silico models have accelerated drug discovery, the predictions made by these models are largely confined to a limited chemical space covered by their training set. Anything beyond this domain can be risky. However, by quantifying uncertainty, researchers can understand the reliability and confidence level of predictions. Similarly, despite the high performance of machine learning algorithms for skin lesion classification, real-world applications remain scarce due to the lack of uncertainty quantification in predictions, which may lead to misinterpretations [4].

In another study, [1] offered a thorough survey on OOD detection and highlighted the importance of establishing a unified framework. In the same vein, [5] addressed the poor generalization performance of convolutional neural networks (CNNs) in medical image analysis. They found that CNNs often failed to detect adversarial or OOD samples. By employing a Mahalanobis distance-based confidence score, their work indicated improved model performance and robustness.

Also, [2] critiqued the state of uncertainty quantification methods in deep learning. They argued that while some OOD inputs can be detected with reasonable accuracy, the current approaches are still not wholly reliable for robust OOD detection. This resonates with [6] who conducted a systematic review of uncertainty estimation in medical image classification. They identified Monte-Carlo Dropout and Deep Ensembles as prevalent methods, highlighting the potential of collaborative settings between AI systems and human experts.

## 3.0    Methods

### 3.1    Experiment 1: Classify OOD images with pre-trained models

Images were generated via DALL-E using search terms like "snail wearing a graduation cap, holding a diploma", "a chainsaw made out of flowers and leaves", etc. The ground truth labels for the five images are chainsaw, lion, snail, car, and dam. For the classification task, five pre-trained neural networks were selected; they include ResNet-50, VGG16, Densenet121, Alexnet, and GoogleNet. This selection captures the diversity of architectures, varying depths, and model complexity, while also representing different milestones in deep learning. For instance, AlexNet is an early forerunner in the field, GoogleNet is based on the inception module [7], Resnet pioneered the use of residual connections [8], VGG16 demonstrated the effectiveness of deeper networks (introducing nonlinearities to learn more complex patterns) [9], and Densenet leveraged dense convolutional networks, connecting layers in a *feed-forward fashion* [10]. The five images used for Experiment 1 are shown in Figure 1 below.

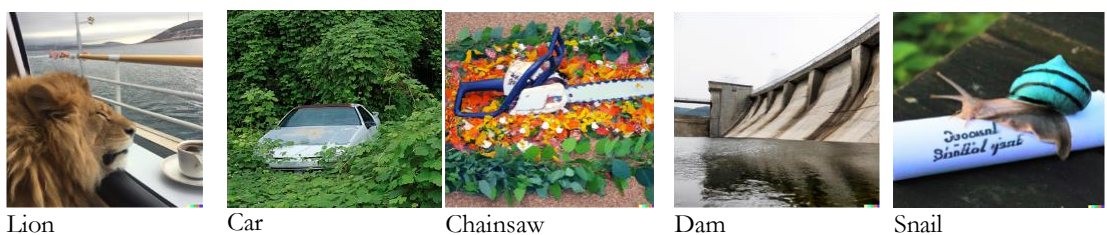

Lion    Car    Chainsaw    Dam    Snail

Figure 1. Out-of-Distribution Images used for Experiment 1.

### 3.2   Experiment 2: Build an uncertainty quantification ensemble

The five neural networks selected earlier served as candidate models for ensemble construction. These members' independent predictions are combined through a committee consensus method. For the consensus method, two options were considered: probabilistic averaging and non-linear combining methods (i.e. majority and plurality voting). The models' predictions (Table 1) for one of the five images in section 3.1 provided a strong rationale for deciding which consensus method to choose.

Table 1. Models' Predictions for Chainsaw

| Model | Prediction |
| --- | --- |
| ResNet-50 | chainsaw |
| VGG16 | wheelbarrow |
| DenseNet121 | wheelbarrow |
| AlexNex | greenhouse |
| GoogleNet | chainsaw |

Here, implementing majority voting for the ensemble would not work, as no class has a clear majority (i.e., >50% of predictions). Plurality voting also proves ineffective, as two classes – chainsaw and wheelbarrow – have equal votes (two each). A possible approach is random selection between the two, but this is synonymous to an 'uninformed gambling' and undermines the principles of responsible AI. A better approach is to obtain the confidence scores of the predictions (e.g., probabilities). Therefore, given that there are cases where majority or plurality voting may fail and require the use of probabilities, isn't it more efficient to directly implement probabilistic averaging? This reasoning forms the basis for selecting probabilistic averaging as the consensus method for the ensemble.

### 3.3   Experiment 3: Test the robustness of best performing model

Four new epistemic images were generated via DALL-E, while one (cat) was photographed in real life. Some perturbations were then added to the images e.g. image rotation, filters, etc. The original and perturbed images were passed through ResNet-50, being the best performing model.

## 4.0   Results and Discussion

### 4.1   Experiment 1

The analysed models recorded different accuracy in classifying the test images. ResNet-50 correctly predicted all five images, DenseNet121 and GoogleNet each correctly classified four out of five, VGG16 classified three images accurately, and AlexNet only managed to correctly classify one image. The specific predictions made by each model is presented in Table 2. ResNet-50's superior accuracy may be linked to its residual connections, which improve gradient flow and feature learning [8]. Though VGG16, DenseNet121, and GoogleNet have varying depths and complexities, they record similar performance highlighting that deeper connections and inception modules may be playing key roles. The poor performance of AlexNet underscores the advancement in deep learning, with more recent neural networks exhibiting better generalization capabilities. Interestingly, the models' performance is consistent with their ImageNet accuracies: ResNet-50 (79.41%), DenseNet121 (74.98%), VGG16 (74.4%), and AlexNet (63.3%) [11]. In a nutshell, ResNet-50, with the highest ImageNet accuracy, is the only model to classify all test images correctly, while AlexNet, with the lowest accuracy, produced the least number of correct predictions.

Table 2. Classification of the five selected images by each model

| Model | Chainsaw | Lion | Snail | Car | Dam |
|---|---|---|---|---|---|
| ResNet-50 | chainsaw | Lion | snail | station wagon | dam |
| VGG16 | wheelbarrow | Lion | slug | station wagon | dam |
| DenseNet121 | wheelbarrow | Lion | snail | station wagon | dam |
| AlexNet | greenhouse | chow chow | mousetrap | Tent | dam |
| GoogleNet | chainsaw | chow chow | snail | station wagon | dam |

## 4.2 Experiment 2

The ensemble model classified all five images correctly, demonstrating its superiority over using the models individually. The average probabilities, variance, and entropy scores for each image were obtained to measure the ensemble uncertainty. Table 3 presents a ranking of the images, starting with the one for which the ensemble model exhibits the highest uncertainty. The ensemble model displayed its highest level of uncertainty when classifying the 'Snail' image, having an average probability of 0.223893, a variance score of 0.149749, and an entropy score of 4.408561. In contrast, the ensemble model showed the least uncertainty when classifying the 'Dam' image, with an average probability of 0.995, a variance score of 0.000045, and an entropy score of 0.043793. For comparison, the maximum entropy score for the ImageNet dataset with 1000 classes is 9.7. That is, the closer the entropy score is to 9.7, the higher the ensemble uncertainty. In the rare case of an ensemble model being 100% certain, the entropy score would be 0.

Table 3. Ranking of images starting with one with most ensemble uncertainty

| Ground Truth | Ensemble | Avg Probability | Variance | Entropy |
|---|---|---|---|---|
| Snail | Snail | 0.223893 | 0.149749 | **4.408561** |
| Car | station wagon | 0.340443 | 0.101638 | 3.306526 |
| Lion | Lion | 0.416775 | 0.277496 | 2.781448 |
| Chainsaw | Chainsaw | 0.423301 | 0.316176 | 2.560379 |
| Dam | Dam | 0.995382 | 0.000045 | **0.043793** |

## 4.3 Experiment 3

Table 4 presents the classification results obtained before and after perturbation was added to the images in Experiment 3. The teddy bear image, rotated by 180 degrees, was misclassified as a cowboy hat. It is likely that the model is interpreting the skateboard (that the teddy was standing on in the original image) as a hat. Also, adding a filter to the cat image led the model to misclassify it as a bucket. Meanwhile, a human is unlikely to make such a fatal error. The snail image, rotated 180 degrees, was misidentified as chocolate syrup. For this, it is possible that the model misinterpreted the snail and its tentacles as flowing chocolate. In addition, the model likely focused on the chainsaw's engine only when it misclassified the chainsaw image as a padlock.

These experiments show that the model relies on some patterns or features in images to make classification and when such features are obscured or altered, the model struggles to make accurate classification.

Table 4. ResNet-50's classification results before and after perturbation

| Ground Truth | Original class | Perturbation added | Perturbed class |
|---|---|---|---|
| Cat | tabby cat | Filter | Bucket |
| Chainsaw | chainsaw | Filter | padlock |
| Teddy bear | teddy bear | Rotation | cowboy hat |
| Lion | Lion | Rotation | Mask |
| Snail | rhinoceros beetle | Rotation | chocolate syrup |

## 4.4 Analysis of saliency maps for the images

Saliency methods offer valuable visual explanations on the inner workings of neural networks by identifying the most critical parts of an input image that contribute to a model's classification decision, thereby improving model's interpretability. These methods can be gradient-based (e.g. SmoothGrad, Vanilla Grad, GradCAM, etc.) or occlusion- and perturbation-based. The SmoothGrad method was selected for this study. With the saliency maps, we obtained additional helpful information in understanding the part of the images considered by the model during classification. For example, the lion image's saliency map showed that the model gets the classification right by focusing on the lion itself, while ignoring the water, ship, and coffee in the image. In contrast, the chainsaw image was misclassified as a padlock, with the map revealing that the model concentrated mainly on the chainsaw's engine, neglecting the saw. When examining the cat images, the saliency maps indicated that the model correctly classified the original image by focusing on the entire cat. However, in the perturbed image, the model's focus shifted to only a portion of the cat's body, leading to a misclassification as a bucket. Lastly, with the snail wearing a graduation hat, the model incorrectly identified it as a rhinoceros beetle. The saliency map shows that the model might have mistaken the hat's edge as a beetle's horn or hind leg.

The saliency maps for the Ensemble mode's classification of images (in Experiment 2) are presented in Figure 2.

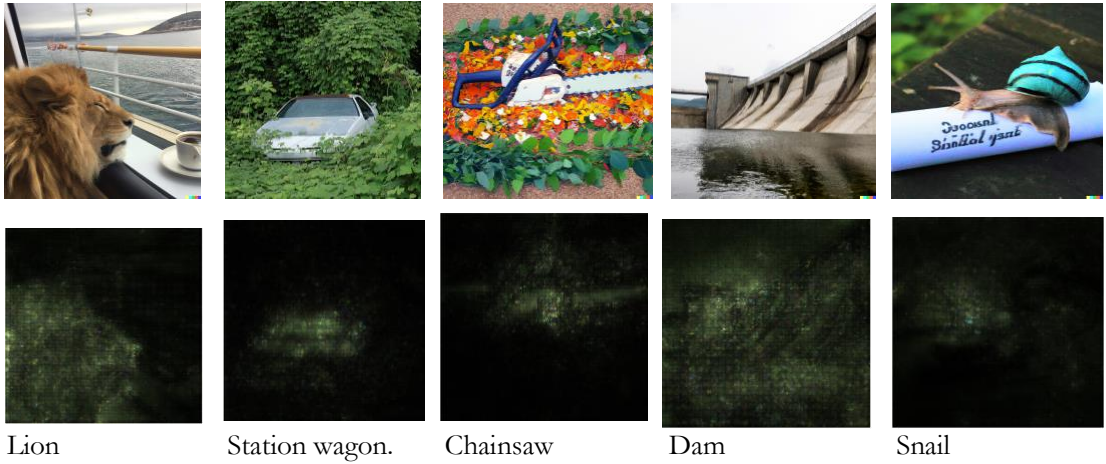

| Lion | Station wagon. | Chainsaw | Dam | Snail |

Figure 2. Saliency maps for Ensemble's classification (Experiment 2)

The saliency maps for the ResNet-50's classification of images (in Experiment 3) are presented in Figure 3. The model's classification of the original and the perturbed images are stated under each image.

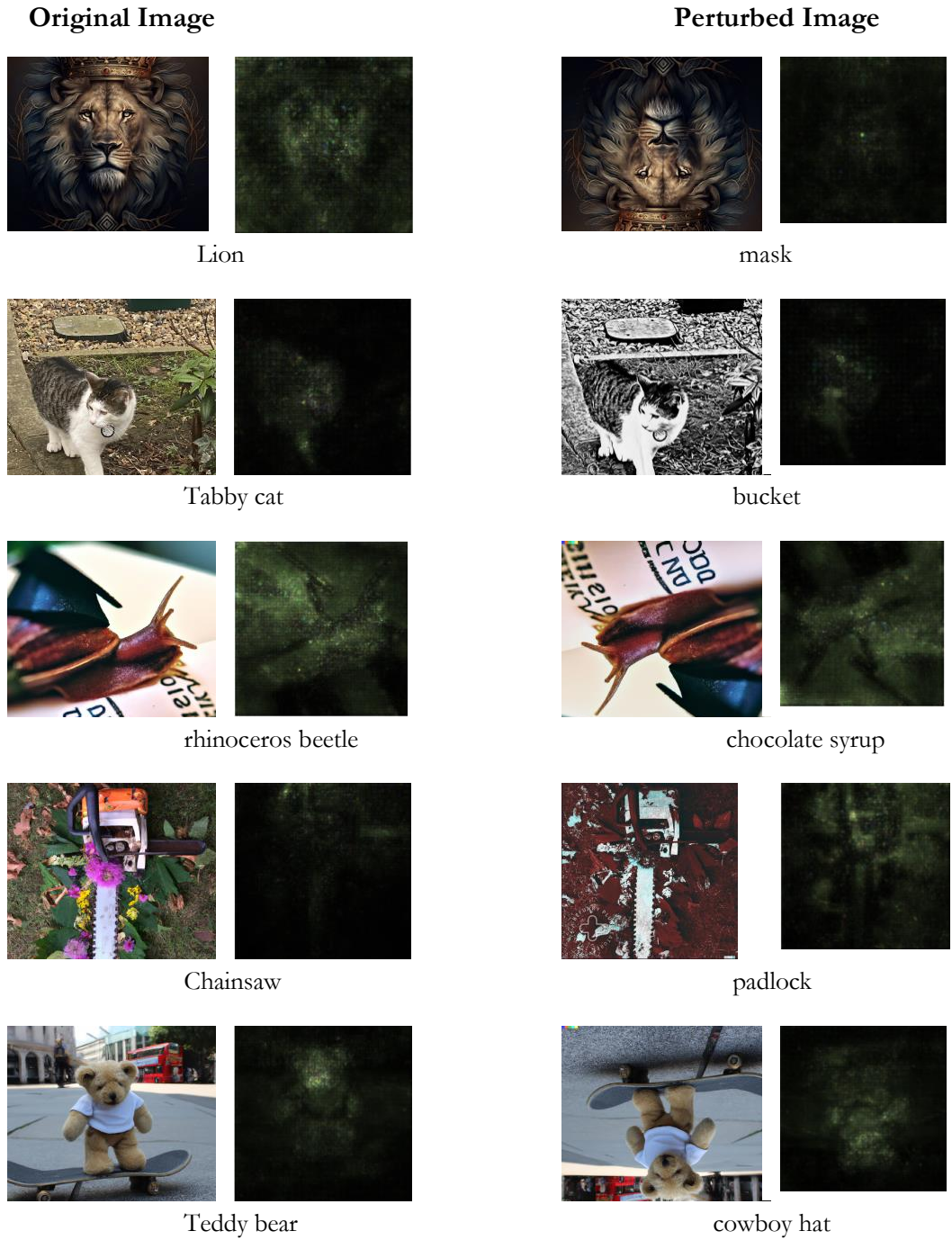

Figure 3. Saliency maps for ResNet-50's classification before and after perturbation

## 5.0   Conclusion and Future Research

Image classifiers can be an enabler of the SDGs e.g. detecting tumors in medical images, analyzing satellite imagery for flood assessment, facial recognition for security systems, etc [12, 13]. However, they could also act as inhibitors to the SDGs e.g. bias against certain groups (Google algorithm classifying African-American men as gorillas [14]), generation of DeepFake images, and vulnerability to adversarial attacks. With minor perturbations, the RestNet50 model in this study made a 180-degree turn to misclassify chainsaw as padlock and cat as bucket. Therefore, it is critical to build image classifiers that are robust to adversarial attacks or distribution shifts. Aside from adversarial training, some strategies are emerging in literature e.g. defensive distillation, gradient masking, Bayesian uncertainty estimation, and feature squeezing [15]. Future research should consider incorporating some of these strategies to improve model robustness in the face of out-of-distribution data.